\documentclass[10pt,twocolumn,letterpaper]{article}

\usepackage{wacv}
\usepackage{times}
\usepackage{epsfig}
\usepackage{graphicx}
\usepackage{amsmath}
\usepackage{amssymb}

\usepackage{subcaption}
\usepackage{array}
\usepackage{tabu}
\usepackage[english]{babel}
\usepackage[utf8]{inputenc}
\usepackage{algorithm}
\usepackage[noend]{algpseudocode}
\usepackage{caption}
\usepackage[labelsep=colon]{caption}
\usepackage{mathtools}
\usepackage{bm}

\usepackage{multirow}

\wacvfinalcopy 

\def\wacvPaperID{****} 

\ifwacvfinal\fi
\begin{document}

\title{Adversarial Code Learning for Image Generation}

\author{Jiangbo Yuan\\
{\tt\small jiayuan@ebay.com}
\and
Bing Wu \\
{\tt\small wubing@sensetime.com}
\and
Wanying Ding \\
{\tt\small wanying.ding@jpmchase.com}
\and
Qing Ping \\
{\tt\small pingqing@amazon.com}
\and
Zhendong Yu \\
{\tt\small ggfish@gmail.com}
}

\maketitle

\begin{abstract}
We introduce the “adversarial code learning” (ACL) module that improves overall image generation performance to several types of deep models. Instead of performing a posterior distribution modeling in the pixel spaces of generators, ACLs aim to jointly learn a latent code with another image encoder/inference net, with a prior noise as its input. We conduct the learning in an adversarial learning process, which bears a close resemblance to the original GAN but again shifts the learning from image spaces to prior and latent code spaces. ACL is a portable module that brings up much more flexibility and possibilities in generative model designs. First, it allows flexibility to convert non-generative models like Autoencoders and standard classification models to decent generative models. Second, it enhances existing GANs’ performance by generating meaningful codes and images from any part of the \textit{priori}. We have incorporated our ACL module with the aforementioned frameworks and have performed experiments on synthetic, MNIST, CIFAR-10, and CelebA datasets. Our models have achieved significant improvements which demonstrated the generality for image generation tasks\footnote{This work was partially done when the authors worked at Vipshop (US).}.
\end{abstract}

\section{Introduction}
The ease with which we design models to generate visually natural and meaningful images, identifies the capability of a machine we built to recognize the real world, let the generative model learning become a rapidly advancing research topic \cite{goodfellow2014generative, radford2015unsupervised, salimans2016improved, arjovsky2017wasserstein, denton2015deep, mirza2014conditional, zhu2017unpaired, denton2015deep}. The community has recently made impressive progress, in particular after Goodfellow et al. proposed the inspiring Generative Adversarial Network (GAN) \cite{goodfellow2014generative}. GAN models often generate much sharper images than autoencoder models such as Deep Autoencoder (AE) \cite{hinton2006reducing}, Denoising AE \cite{vincent2010stacked}, Variational AE \cite{kingma2013auto}, Adversarial AE \cite{makhzani2015adversarial} and so on. PixelRNNs \cite{dahl2017pixel} represents an alternative model family that can produce equally or even sharper images than GANs. These recent advances of generative modeling have shown huge potential to many computer vision tasks, e.g. image inpainting \cite{zhu2017unpaired,pathak2016context,yeh2017semantic,yan2016attribute2image}, semi-supervised learning \cite{dai2017good, salimans2016improved,lecouat2018semi, belghazi2018hierarchical, springenberg2015unsupervised,belghazi2018hierarchical,balestriero2017semi}, data manipulation \cite{brock2016neural, zhu2016generative}, high-resolution image generation \cite{ledig2017photo, karras2017progressive, sonderby2016amortised, parmar2018image}, transfer learning \cite{bousmalis2017unsupervised, tzeng2017adversarial, hoffman2017cycada,sankaranarayanan2017generate}, image-to-image translation
\cite{zhu2017unpaired,yi2017dualgan, liu2017unsupervised, choi2017stargan, zhu2017toward, isola2017image}, and text-to-image \cite{reed2016generative, reed2016learning, zhang2017stackgan, xu2017attngan, zhang2018photographic}, to name a few. 
\begin{figure}[t]
\centering
  \includegraphics[trim={1cm 0 4cm 0}, clip, height=0.45\linewidth,]{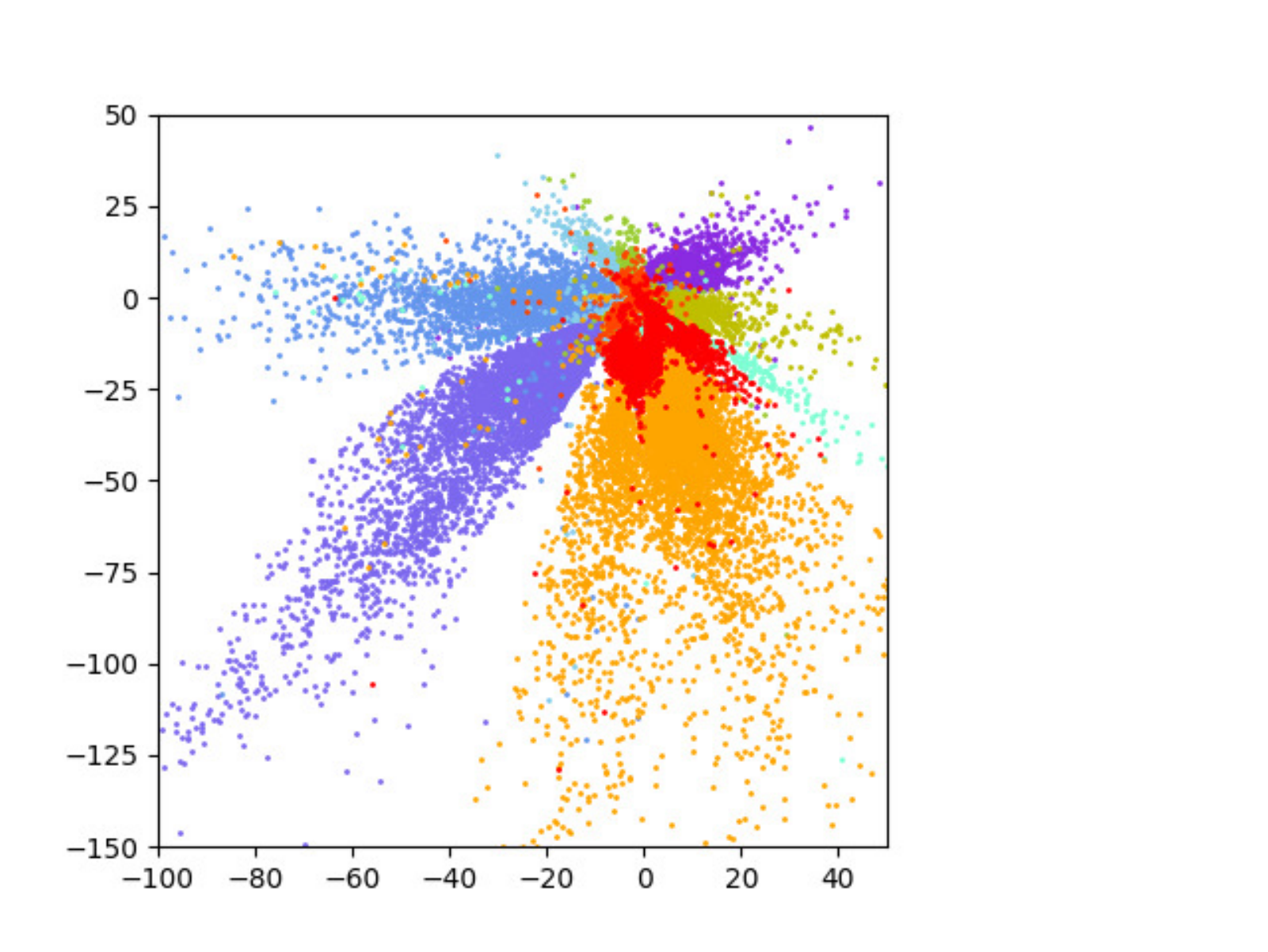}
  \includegraphics[trim={1cm 0 1cm 0}, clip, height=0.45\linewidth]{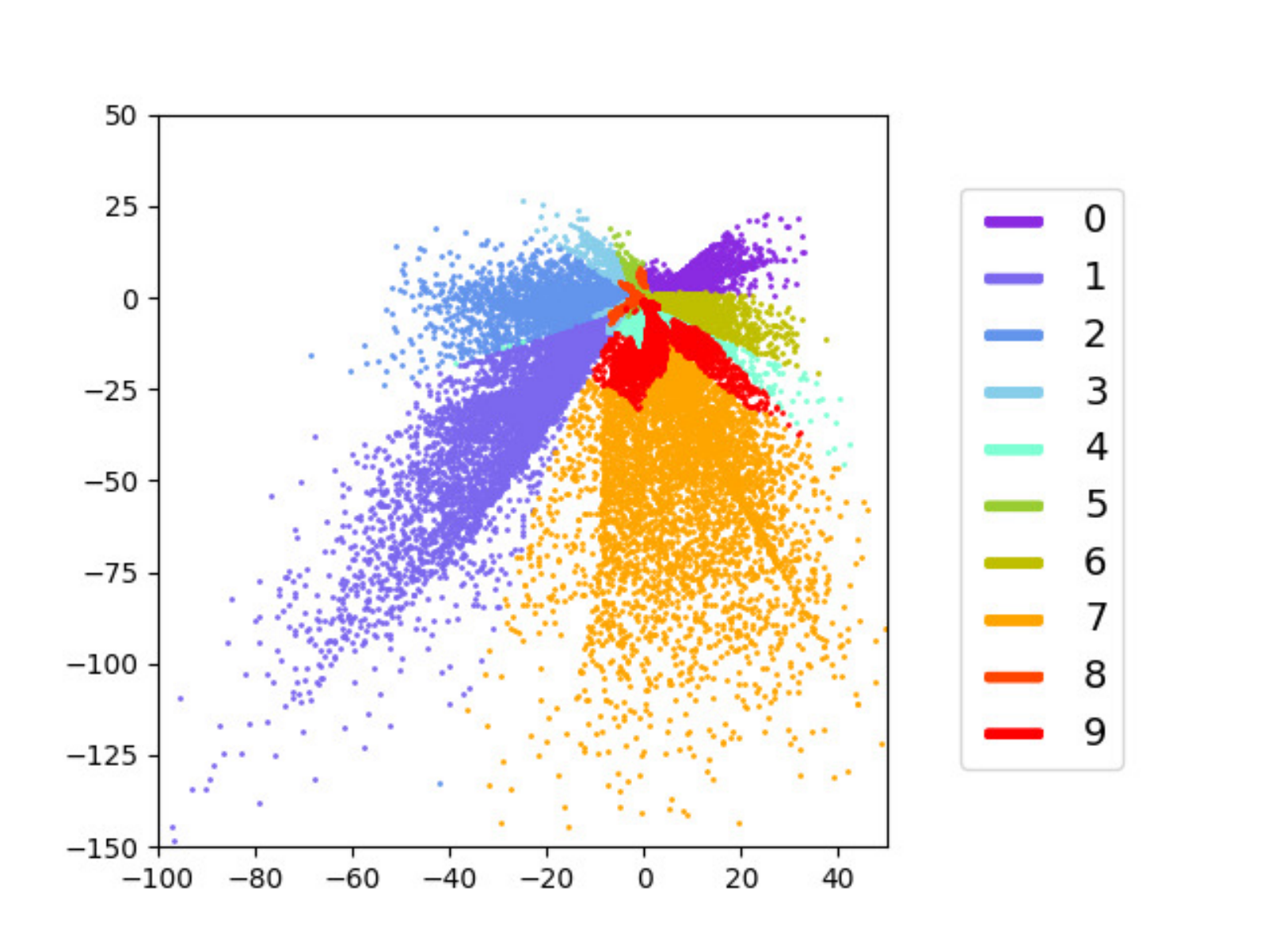}
\vspace{-1em}
\caption{Illustration of the two latent code distributions of our ACL model on MNIST dataset. \textbf{Left} is the learning target, which is the output code distribution of an image encoder, and \textbf{Right} is the learned code distribution that using random noise as its input. The colors represent the 10 digit classes. With a joint learning process, both codes are expected to draw plausible images via an image generator.}
\label{fig:mnist_acl}
\end{figure}

GANs are currently the most popular generative modeling approach considering model output quality, training costs, and generality. While the standard GANs \cite{goodfellow2014generative, salimans2016improved} use powerful adversarial objectives to map some random input noise directly to image-level distributions. Some severe challenges remain. On one hand, since the loss only measures how well the generator fools the discriminator rather than measuring true image quality, a biased discriminator may introduce uncertainties to the learning process. Mode collapse is the most famous one. On the other hand, the implicit latent space learning makes it challenging for code manipulation and latent space interpolation at the inference stage. Follow-up optimization strategies cover almost all model components. For example, searching for alternative objective functions \cite{arjovsky2017wasserstein,berthelot2017began,mao2017least,zhao2016energy} and gradient optimization strategies \cite{gulrajani2017improved,nagarajan2017gradient,roth2017stabilizing} sometimes help to improve the stability of GAN training. Meanwhile, some other researchers try to improve intermediate-level feature matching \cite{salimans2016improved,mroueh2017mcgan} to ease the distribution learning. Some share similar spirits but go one step further focusing on the input layer learning that introduces auxiliary encoders \cite{makhzani2015adversarial,larsen2015autoencoding,dumoulin2016adversarially} or label distributions \cite{mirza2014conditional, perarnau2016invertible,bao2017cvae} to regularize the input code distribution for GANs. Our work falls into this type of optimization strategies, which aims to improve generative modeling by performing a prior code learning. 

The prior code learning is critical because it is the source to the rest of a generative model. Improper input code distributions will propagate and potentially hurt the model performance. There are many possible methods to perform the alignment between a prior distribution and a reference code distribution. The solutions we end up with might be intricate and highly entangled. We have seen, for instances, using a specific type of prior distributions \cite{larsen2015autoencoding,mescheder2017adversarial}, or learning to transform the latent code distributions to some random prior distributions \cite{makhzani2015adversarial}, and so on. Most of the compromises or assumptions result in trade-offs between the model performance and its flexibility. To this end, we present an approach that maps random noise to a reference code space that is regulated by other models. Our approach neither takes assumptions to the prior distributions nor collapses the latent code distributions. Figure \ref{fig:mnist_acl} demonstrates an example of such learned codes in our code learning framework. The adversarial learning maximize the benefits of learning from an inference net, in which a posterior distribution $p(c|x)$ is drawn from the real data $x$. Our contributions are three folds:
\begin{itemize}
  \item We propose a simple yet effective prior code learning approach, which maps a random prior noise to a reference code space; The reference space, for example, could be the bottleneck layer from an Autoencoder, or a feature layer from a discriminative model.
  \item We show that different GAN architectures plugged with the proposed ACL can produce images of better visual quality as well as FID scores \cite{heusel2017gans}.
  \item We demonstrate that ACL is a general portable code learning approach. It provides a way to incorporate standard classification models with generative modeling, where the joint learning retains good property regarding generating images with good quality. 
\end{itemize}

\section{Related Works}

A generative model is of a way to learn data distributions from a set of training data to generate new data points. The most widely used deep generative models are Variational Autoencoder(VAE) \cite{kingma2013auto}, Generative Adversarial Networks (GAN) \cite{goodfellow2014generative}, and a variety of GAN variants such as Wasserstein-GAN(WGAN) \cite{arjovsky2017wasserstein} and WGAN-GP \cite{gulrajani2017improved}.  VAE and its derived models learn a low-dimensional latent representation $z$ from training data and feed $z$ to decoder to generate new data points. Since VAE is optimizing the lower variational bound, the quality of the generated image is somewhat poor compared to GANs. GAN and its derived models are comprised of a discriminator network against a generator network to reach Nash equilibrium. Training GAN sometimes is hard. Models may never converge, and mode collapse are common. WGAN \cite{arjovsky2017wasserstein} improves GAN's performance by replacing the asymmetric KL-divergence in GAN with the Wasserstein Distance. WGAN-GP further applies the Lipschitz constraint to WGAN and achieved more stable and impressive results. 

\noindent \textbf{Latent Code Learning} An unconstrained input to GANs may introduce uncertainties to the training \cite{brock2018large}. A variety of GANs have contributed to an improvement of generator learning by introducing an encoder to GAN framework \cite{larsen2015autoencoding, benaim2017one}. This leads to new hybrid GAN models, i.e. Autoencoding GANs. If we do not explicitly build a relationship between the encoder output (posterior $p(\bm{z}|\bm{x})$) and the input marginal distribution $p(\textbf{\emph{z}})$, the encoder is merely used as a regularization to the generator training in such hybrid models. For example, DistGAN \cite{tran2018dist} use pairwise distances between encoder output codes and random input noise as a constraint to regularize the feature distributions in the last layer of the Discriminator.

It is also intuitive to directly align the two distributions in some ways while fixing the generative learning. AAE \cite{makhzani2015adversarial}, for instance, is a recent such work targeting to map $p(\bm{z}|\bm{x})$ to $p(\bm{z})$. It transforms the autoencoder to a generative model by collapsing the posterior $p(\bm{z}|\bm{x})$ to the random noise $p(\bm{z})$. AAE bears close to our ACL approach w.r.t. the explicit and adversarial learning of the code distribution alignment. The two approaches are, however, exactly reverse to each other due to the opposite mapping directions. Moreover, it is infeasible to use AAE to enhance GANs since the joint input source is supposed to be close to the original input noise.

Brock, et al., propose a large-scale GAN model named BigGAN \cite{brock2018large}, which achieved state-of-the-art FID or IS scores on specific benchmark datasets. Although BigGAN is trained in a supervised setting, the core technique used is called "truncation trick". What the trick does is basically to re-sample the input noise data when it is out of a right "range". BigGAN ultimately uses a prior that is picked from a random distribution. It is worth to mention that the authors use spectral norms to detect an abnormality in the GAN generator layers. They point out that the first layer of the generator is the most vulnerable to trigger mode collapse. This aligns well to our standpoint that the unconstrained input noise makes GANs more unstable. 

Among other works, adversarial discriminative domain adaptation (ADDA) \cite{tzeng2017adversarial} and adversarially regularized autoencoders (ARAE) \cite{junbo2017arae} are close to our method w.r.t. the high-level ideas of adversarial latent distribution learning. However, these methods are conceptually very different. ADDA uses adversarial latent code learning in a supervised setting. It considers the alignment between two domains such as two closely distributed image datasets. However, we merely have one domain from real data distribution, and an arbitrary noise for another. The difference is evident in that the random noise in our problem lies in unlimited space. This sets their method to a transfer learning solution while ours to an adversarial code learning approach. On the other hand, ARAE is a setting for NLP problem and random noise is involved as the input. Our focus is about image generation and the most important assumption in our code learning is that inference nets can provide meaningful latent code distributions for the image generators. Thus we set such latent codes as the learning target instead of mutually aligning both the code and the prior as does in ARAE. We re-implemented ARAE in our image generation scenarios. One can find experimental comparison in Section \ref{exp-comparison-gans}.



\section{The Approach}
Our work belongs to the domain of \emph{Deep Generative Modeling}. In recent years, there has been a resurgence of interests in deep generative models. The capability of a deep generative model is highly associated with its learning ability about the abstract representations. To images, for example, it is comfortable to say that the more satisfying image quality a model achieves, the better latent feature representations are captured by the model, and vice versa. To this end, we desire a type of learnable generative model input component, which is often the most abstract component.  

\subsection{Code Learning in Generative Models}
In deep generative modeling, one does not need to define a probability distribution explicitly but rather train a generative network to draw samples $p_x(\bm{x})$ from some input data. In standard GAN, the noise data fed into the generator are from arbitrary but fixed distribution. In VAE, the input data to the decoder are learned latent vectors that roughly follow a Gaussian distribution. We here generalize such latent variable learning to be a \emph{Code Learning} framework in deep generative modeling. 



Our model will be formulated in a basic encoding-decoding structure. We define a function $E(\bm{x}; \theta_{E})$ to encode data $\bm{x}$ to latent code vectors $\bm{c} \in \mathbb{R}_{d_{c}}$. We then consider $p_{c}(\bm{c})$ as our real code distribution. To learn the decoder distribution ${p}_{G}(\bm{x})$ over data $\bm{x}$, we define a mapping function $G(\bm{c}; \theta_{G})$ that draws samples $\bm{x}$ from $\bm{c}$ as the input. Because this standard encoding-decoding is not a generative model yet, we introduce another function $C(\bm{z}; \theta_{C})$ to transform the prior random noise $\bm{z}$ to the real latent code space $\bm{c}$. We shall now see that two encoding functions $E(\bm{x}; \theta_{E})$ and $C(\bm{z}; \theta_{C})$ will simultaneously learn two posterior distributions $p(\bm{c}|\bm{x})$ and $p(\bm{c}|\bm{z})$, where the real data and the random \textit{priori} flow together over ${p_{c}(\bm{c})}$. Thus far, there are three functions in our framework, with parameters $\theta_{G}$, $\theta_{E}$, and $\theta_{C}$, respectively. Though other forms can be used, we choose neural networks to represent all three functions. In addition, we let $d_{c} = d_{z} = d$ for simplicity.

\subsection{Adversarial Code Learning}
The core component in the proposed code learning framework is the code generator $C(\bm{z}; \theta_{C})$, which is used to estimate the posterior $p(\bm{c}|\bm{z})$. Other researchers have proposed methods to model the relationship between the two distributions, e.g., learning to transfer the code to the noise \cite{makhzani2015adversarial} or using distance constraints to regularize the code encoding \cite{mescheder2017adversarial}. To our best knowledge, we are the first to define a function to map prior noise data to learned code variables explicitly.

\begin{figure}[h]
\begin{center}
  \includegraphics[trim={0 0 0 1.5cm}, clip, width=1\linewidth]{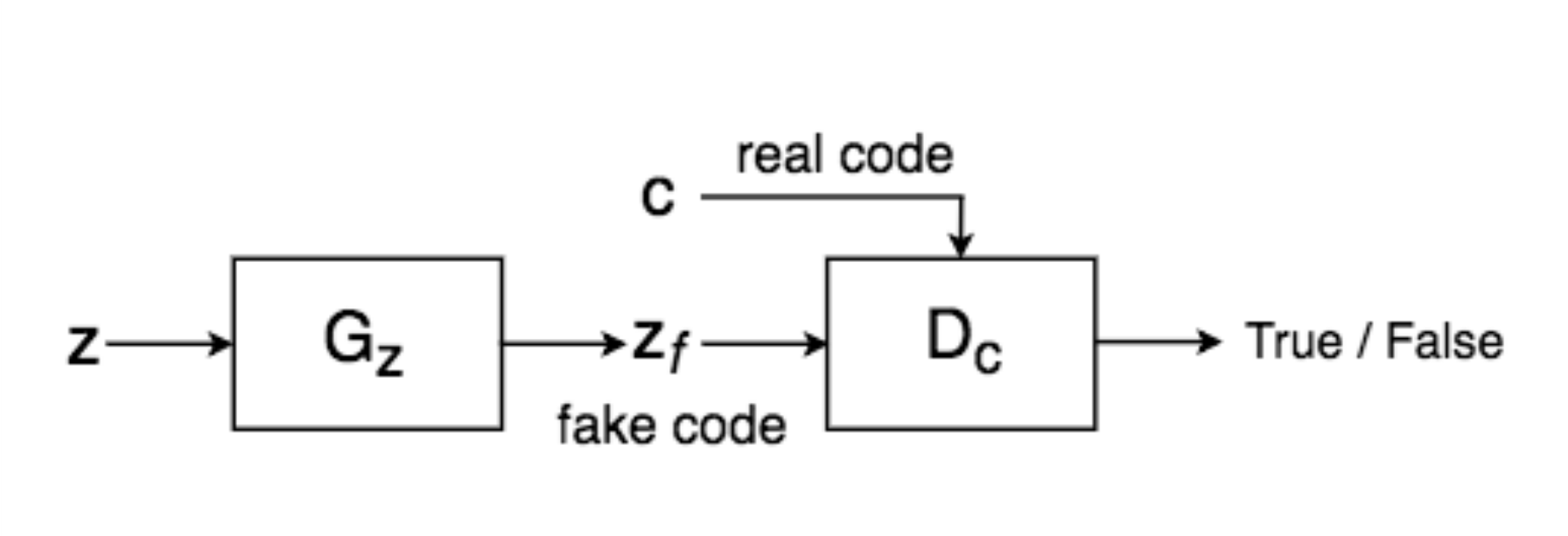}
\end{center}
\vspace{-2em}
\caption{The architecture of the ACL Module.}
\label{fig:acl_module}
\end{figure}

We aim at matching an arbitrary noise data $p_{z}(\bm{z})$ to the learned latent code $p_{c}(\bm{c})$. Note that, here we do not have any assumptions about the distributions of $p_{c}(\bm{c})$. The distribution of $p_{c}(\bm{c})$ is correspondingly determined by the associated objectives in different implementations. The overall structure of the ACL module is illustrated in Figure \ref{fig:acl_module}. The formulations of other components will be discussed in next section of Plugging to Generative Models.

\begin{figure*}[htb]
\centering
    \begin{subfigure}[b]{0.32\textwidth}
      \centering\includegraphics[trim={0.5cm 0.3cm 0.5cm 0.3cm}, clip, width=1\linewidth]{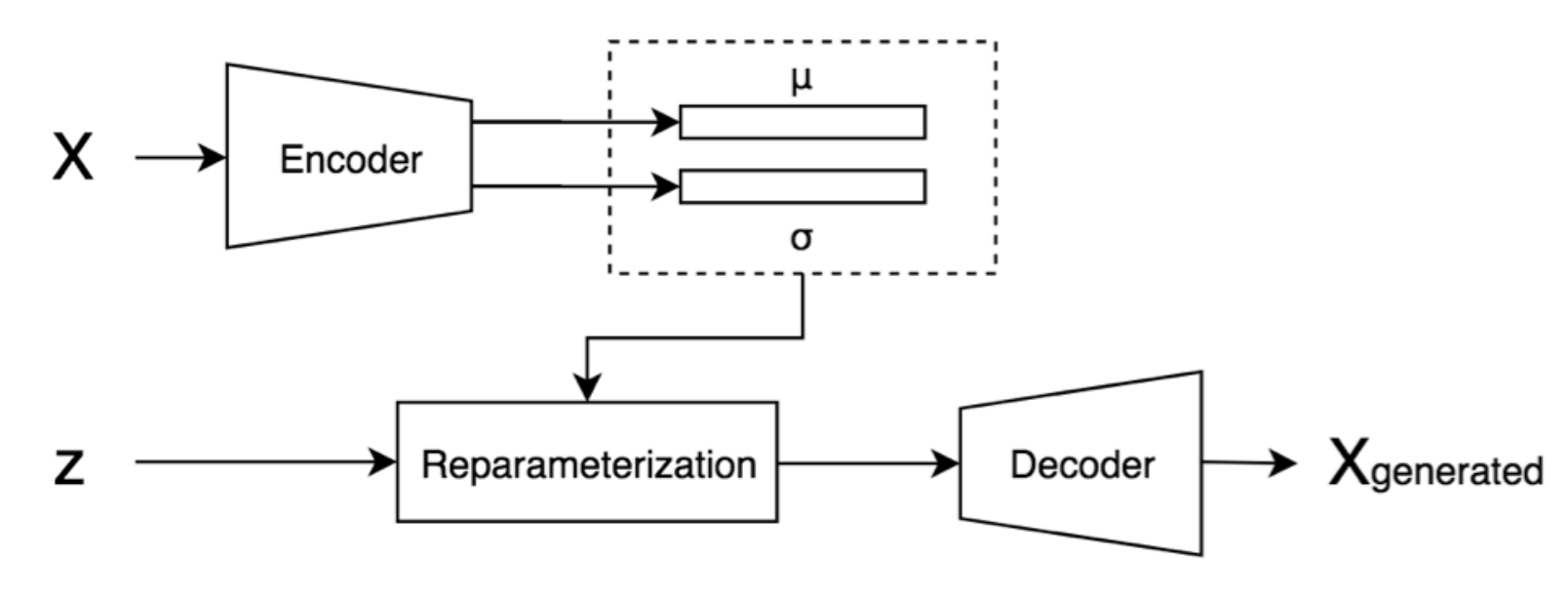}\caption{}
    \end{subfigure}
    \begin{subfigure}[b]{0.32\textwidth}
      \centering\includegraphics[trim={0.5cm 0.3cm 0.5cm 0.3cm}, clip, width=1\linewidth]{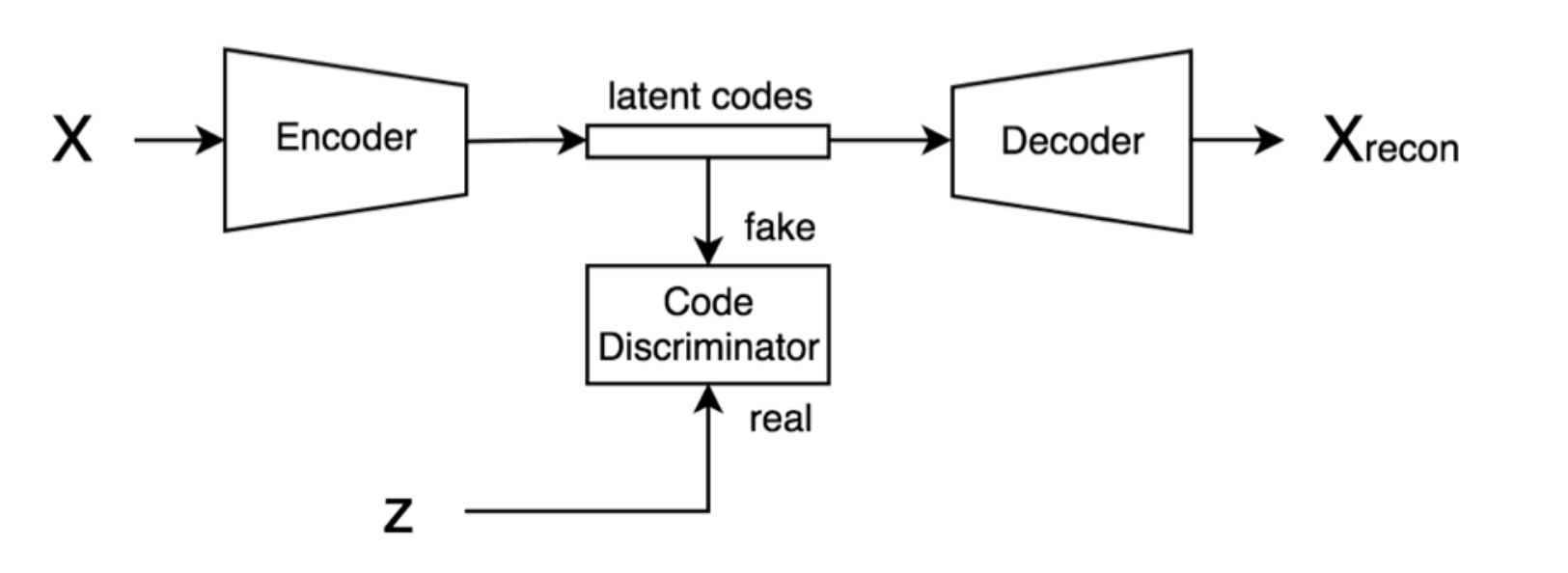}\caption{}
    \end{subfigure}
    \begin{subfigure}[b]{0.32\textwidth}
      \centering\includegraphics[trim={0.5cm 0.3cm 0.5cm 0.3cm}, clip, width=1\linewidth]{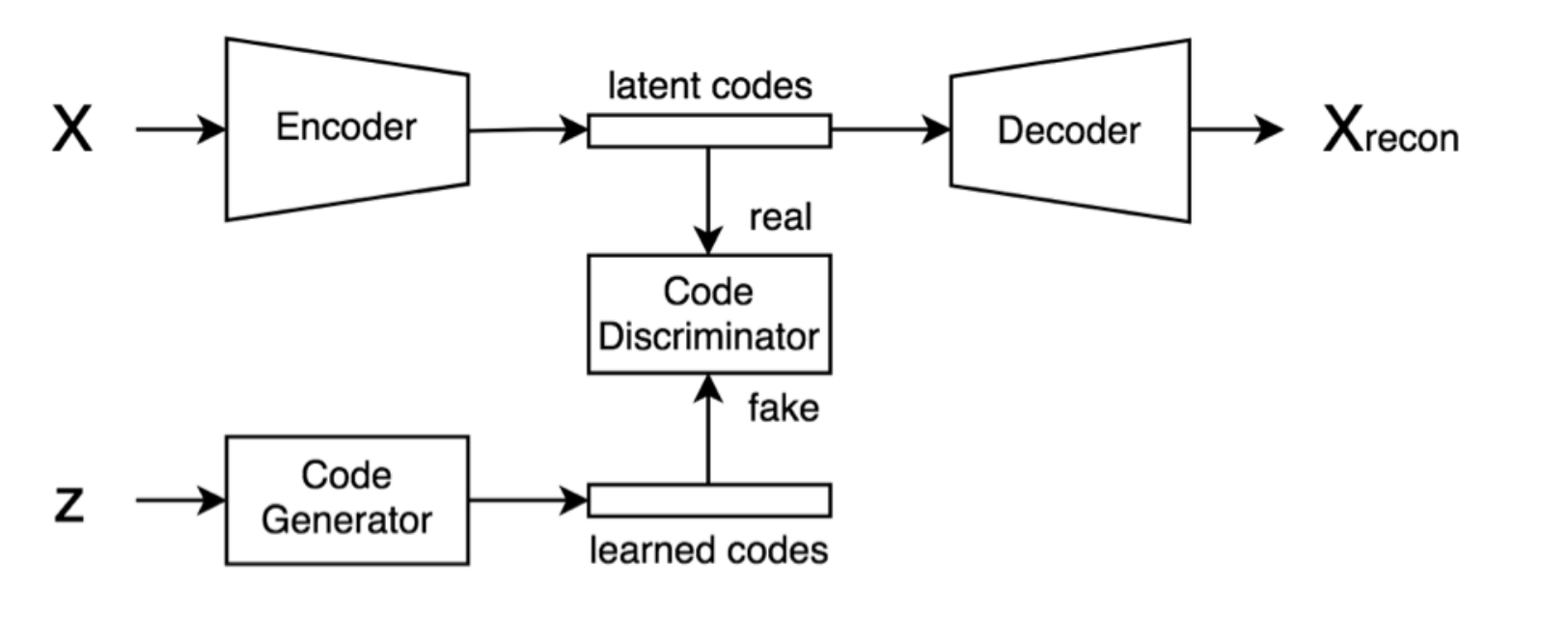}\caption{}
    \end{subfigure}
    \vspace{-1em}
\caption{(a) Variational Autoencoder; (b) Adversarial Autoencoder; (c) Our ACL Autoencoder}
\label{fig:ae_arch}
\end{figure*}

Inspired by GANs, which maps the imposed prior of $p_{z}(\bm{z})$ to complex data distributions $p_{x}(\bm{x})$, we introduce a similar adversarial procedure for code learning. We introduce a code discriminator $D_c(\bm{c})$, along with the defined generator $G_z(\bm{z})$, to kick off the adversarial game. The generator $G_{\bm{z}}$ samples $\bm{z}$ from the marginal prior distribution $p_z(\bm{z})$ to the marginal code distribution $p_c(\bm{c})$. $G_z(\bm{z})$ is trained to maximally confuse the discriminator $D_c(\bm{c})$ to draw equal probabilities over the generated codes and the true latent codes, towards the end of the learning. Note that, we are discussing only the code learning phase in the whole framework, where we assume the marginal distribution $p_c(c)$ is fixed to the code learner. The solution to this game can be expressed as an adversarial loss as follows: 
\begin{multline}
\mathcal{L}_Z = \underset{G_{z}}{min} \underset{D_{c}}{max} E_{c \sim p_{c}} \left [ log\mathit{D_{c}(\bm{c})} \right ] +\\
E_{z \sim p_{z}} \left [ log (1-\mathit{D_{c} (G_{z}(\bm{z}))}) \right ]
\label{eqn:GAN_code}
\end{multline}

It would be identical to the original GAN's objective if we replace $\bm{c}$ with $\bm{x}$. They are yet conceptually different. Original GANs set the target to sampling the raw data distribution $p_{x}$ while ACL aims to draw codes from a given latent code distribution $p_{c}$. The former is at the output layer, and the latter is at the input layer, in their corresponding deep generative modeling frameworks. Due to the inherent adversarial nature shared by the two, most optimization strategies for GANs are intuitively applicable to our ACL. However, we focus on simple implementations of ACL throughout this paper.


\section{Generative Models with ACL}

Throughout this section, we concern with different generative models that our ACL models can be easily plugged into, with minimal modifications. As we discussed above, ACL is best seen as a module that can be integrated to existing architectures only if there is a target latent code space. It opens a door of possibilities to transfer non-generative models to generative ones, and to boost existing generative model performance. We here present a few examples to demonstrate the potential usage of ACL.

\subsection{ACL with Autoencoders}
Deep Autoencoder \cite{hinton2006reducing} and its variants \cite{kingma2013auto, makhzani2015adversarial, vincent2010stacked} are popular unsupervised learning models. Its simple encoder-bottleneck-decoder structure projects raw data to low-dimensional latent vectors. With a reconstruction constraint as objective, the Autoencoders can capture underlying data distributions in the latent code space. Autoencoders are not generative models yet in that the learned latent code distribution is not directly accessible. However, they are excellent examples for our ACL integration, which we call ACL-Autoencocders (ACL-AE).

The architecture of ACL-AE is illustrated in Figure \ref{fig:ae_arch}(c). We attach our ACL module to a vanilla Autoencoder (with an encoder $Enc_{x}$ and a decoder $Dec_{x}$). The bridge of AE and ACL is the adversarial code learning objective as in Equation \ref{eqn:GAN_code}, that pushes the generated latent codes to be as close to the latent codes from the encoder.

The Autoencoder objective can be expressed as: 
\begin{equation}
\mathcal{L}_E = \underset{\theta_{Enc},\theta_{Dec}}{min} ||x - Dec_{x}(Enc_{x}(x))||_{2}^{2}
\end{equation}
, where $\theta_{Enc},\theta_{Dec}$ are the parameters of the encoder and the decoder, and $\bm{x}$ is the input image. The adversarial objective for latent code learning is as in Equation \ref{eqn:GAN_code}. The overall objective function for ACL-Autoencoder is defined as:
\begin{equation}
\mathcal{L}_{ACL-AE} = \mathcal{L}_E + \lambda_1 \cdot \mathcal{L}_Z
\end{equation}
Here we use the simplest parameter setting of $\lambda_1=1.0$.

To demonstrate how our model learns the mapping to the latent code space differently from existing models, we illustrate the model architecture of VAE, AAE, and our model in Figure \ref{fig:ae_arch} . VAE is designed such that latent code distributions from encoders roughly follow a Gaussian distribution (Figure \ref{fig:ae_arch}(a)). AAE, on the other hand, collapsing the posterior $p(\bm{c}|\bm{x})$ to a fixed and random noise distribution $p(\bm{z})$ (Figure \ref{fig:ae_arch}(b)). AAE can be seen as a reverse to our model. However, it is infeasible to use AAE to enhance GANs since the joint input source is supposed to be close to the original input noise. 

\subsection{ACL with GANs}
\begin{figure}[!htb]
\begin{center}
  \includegraphics[width=1\linewidth]{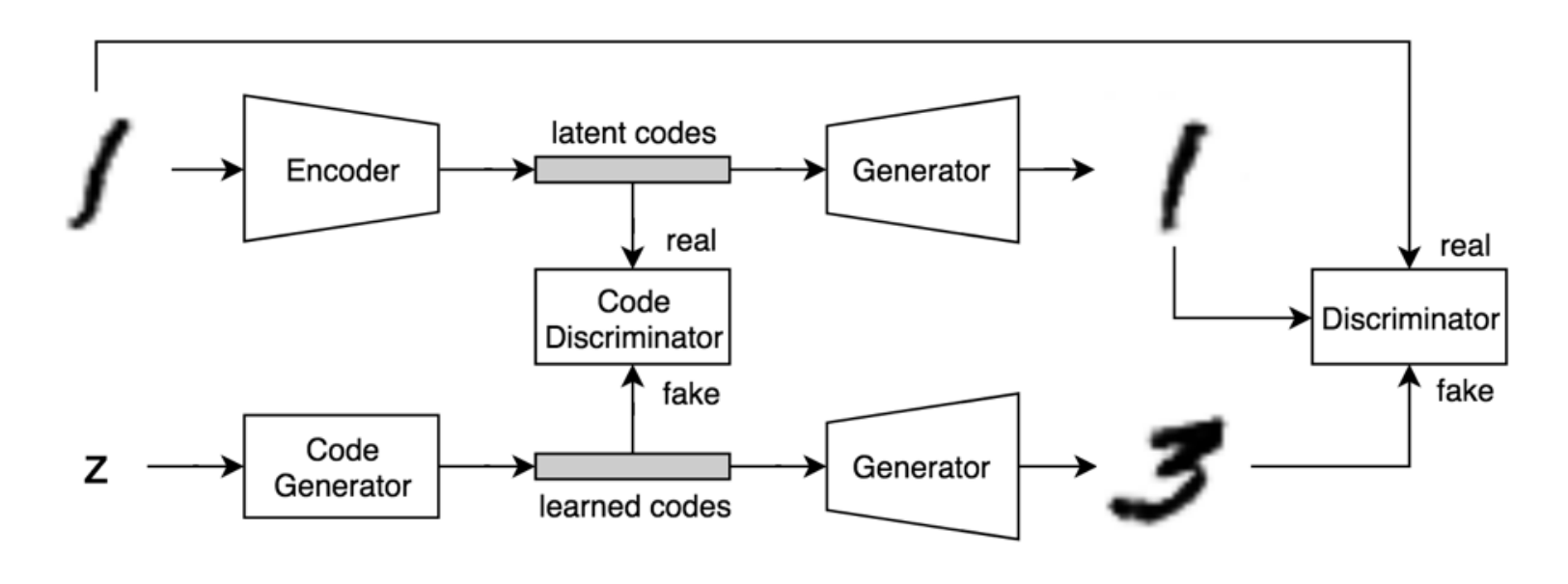}
\end{center}
\vspace{-2em}
\caption{The architecture of ACL-GAN.}
\label{fig:gan_arch}
\end{figure}

 Among various approaches to optimizing the training of GANs, the ones that regularize GANs through encoder networks \cite{larsen2015autoencoding,makhzani2015adversarial} are of particular interest to the present study. Enlightened by this idea, we combine Autoencoder with GAN and plug in our ACL module. We first encode the data samples to latent codes with encoder in a typical Autoencoder framework, and then collapse the decoder of Autoencoder and generator in GAN into one. The ACL is plugged onto the bottleneck part after encoder. The architecture of ACL-GAN is shown in Figure \ref{fig:gan_arch}. The input images are considered as "real" and the generated samples are considered as "fake" for the image discriminator. The output of the image discriminator for the reconstructed samples is only used for training the encoder. The latent code generator and discriminator are the same as in Section 3.2. 
 
 As for the element-wise reconstruction loss in conventional Autoencoders, which often yield blurry images, we replace it with a feature-wise similarity metric between features in the generator and discriminator \cite{larsen2015autoencoding}. The objective of the Autoencoder module therefore becomes: 
\begin{equation}
\mathcal{L}_{rec} = \underset{\theta_{E_x},\theta_{G_x}}{min} ||\Phi(\bm{x}) - \Phi(G_{x}(E_{x}(\bm{x})))||_{2}^{2}
\end{equation}
where $\Phi(\bm{x})$ is the feature output from the last layer of the discriminator.
The adversarial objective for code learning stays the same with Equation \ref{eqn:GAN_code}, and the adversarial objective for image learning is defined as:
\begin{multline}
\mathcal{L}_{GAN} = \underset{\theta_{G_x}}{min} \underset{\theta_{D_x}}{max} E_{x \sim p_{x}} [ log\mathit{D_{x}(\bm{x}))} ] + \\
E_{c \sim p_{c}} [ log (1- \mathit{D_{x}(G_{x}(\bm{c}))}) ]
\label{eqn:GAN}
\end{multline}
where $D_{x}(\bm{x})$ is an image discriminator that computes the probability of $x$ being a sample from the data distribution, and $G_{x}(\bm{c})$ is an image generator that maps a sample $\bm{c}$ from the latent code space to the data space. Different from the original GAN's objective, we are not generating images from an arbitrary prior distribution, but from the learned code distribution.
The overall objective function for our ACL-GAN, is therefore defined as:
\begin{equation}
\mathcal{L}_{ACL-GAN} = \lambda_2 \cdot \mathcal{L}_{rec} + \lambda_3 \cdot \mathcal{L}_{GAN} + \lambda_4 \cdot \mathcal{L}_{Z}
\end{equation}
Here we use the simplest combination of $\lambda_2=\lambda_3=\lambda_4=1.0$. The training process is summarized in Algorithm \ref{alg:aclgan}.

In the experiment section, we also demonstrate the flexibility of ACL, by plugging into one of the state-of-the-art GAN variants, i.e. WGAN-GP \cite{gulrajani2017improved}, by introducing Gradient Penalty into our ACL-GAN framework.

\subsection{ACL with Supervised Learning}
GAN provides a powerful way to generate images, but also comes with costs: without supervision, such models provide no means of exerting control over features to be found. Currently, there are mainly two kinds of supervisions widely in use: labels and prior. IcGAN \cite{perarnau2016invertible}, CGAN \cite{mirza2014conditional}, CVAEGAN \cite{bao2017cvae} explicitly incorporate categorical labels on GANs or VAEGANs. InfoGAN \cite{chen2016infogan}, on the other hand, disentangles representations by maximizing the mutual information between latent variables and observations, in which observations present as prior. 

The works employ supervision information such as real labels or categorical priors to guide the GAN training, which reacts as regularization to the supervised models. That often aims to improve a semi-supervised performance. The findings in BadGAN model \cite{dai2017good} provides a strong indication that such joint GAN models appear to produce better semi-supervised results with subjectively worse images. 

\begin{figure}[!htb]
\begin{center}
  \includegraphics[width=1\linewidth]{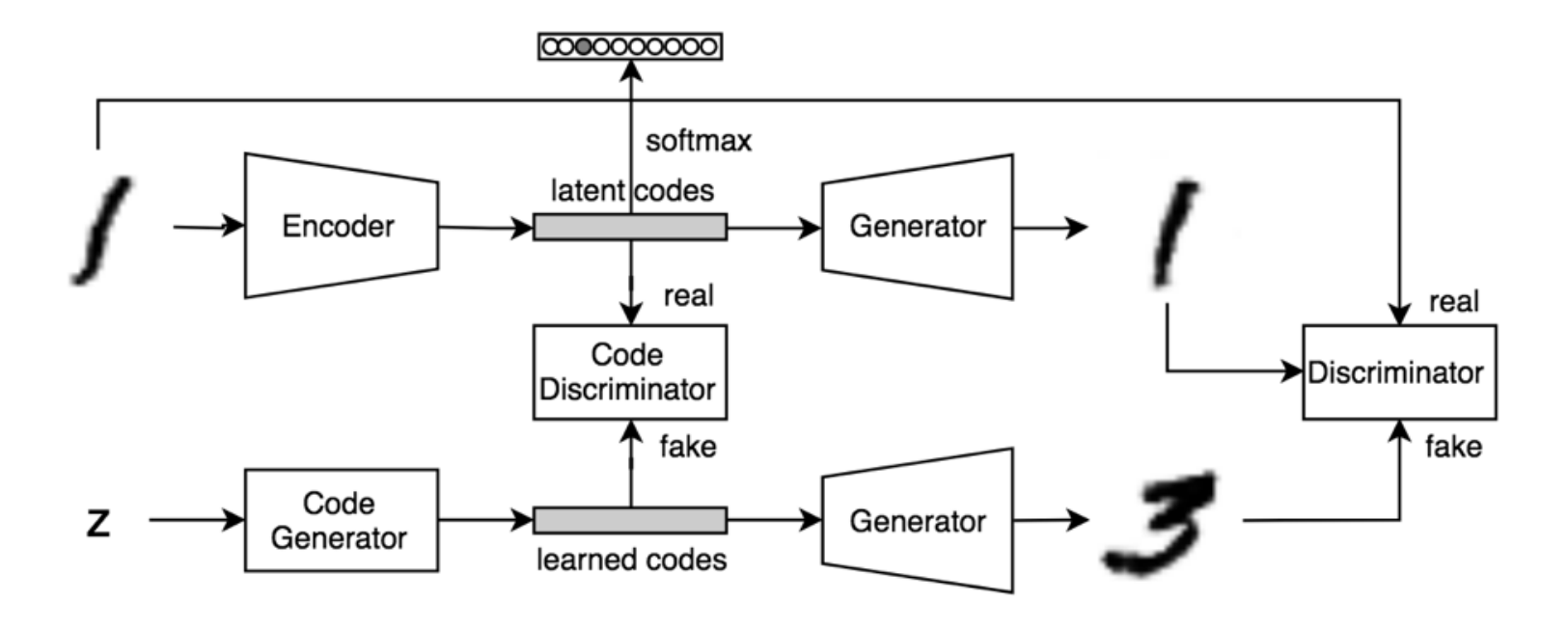}
\end{center}
\vspace{-2em}
\caption{The architecture of ACL-SGAN.}
\label{fig:sgan_arch}
\end{figure}

We here carry on our exploration of the capability of our ACL model, where we set a goal to convert a classic discriminative classification model into a generative one smoothly, and to generate images of good quality. Therefore, we change our inference net to a classification model that is supervised by class labels during the training. As seen in Figure \ref{fig:sgan_arch}, the network structure of this supervised ACL-GAN is very similar to our ACL-GAN model expect that the encoder is replaced by a classifier. We name this model as ACL-SGAN, in which the code learning target becomes the last fully connected layer of the classifier. 

\begin{algorithm}
\caption{The training pipeline of the proposed ACL-GAN }\label{alg:aclgan}
Require: ${\theta}_{E}$, initial image encoder network parameters. ${\theta}_{G_x}$, initial image generator network parameters. ${\theta}_{D_x}$, initial image discriminator network parameters. ${\theta}_{G_z}$, initial latent code generator network parameters. ${\theta}_{D_c}$, initial latent code discriminator network parameters. 
\begin{algorithmic}[1]
\While{${\theta}_G$ has not converged}
\State {Sample ${x_r} \sim p_x$ a batch from real image data; }
\State {$c\leftarrow E_{x}(x_r)$ }
\State {$x_f\leftarrow G_{x}(c)$ }
\State {Sample ${z} \sim p_z$ from random noise distribution; }
\State {$c_f\leftarrow G_{z}(z)$ }
\State {$x_{ff}\leftarrow G_{x}(c_f)$ }

\State $\mathcal{L_D} \sim -(log(D_{x}(x_r))+log(1-D_{x}(x_{ff})))$

\State $\mathcal{L_Z} \sim -(log(D_{c}(c))+log(1-D_{c}(c_f)))$
\State $\mathcal{L_R} \sim ||\Phi(x_r) - \Phi(x_{f})||_{2}^{2}$

\State $ \theta_{E_x} \stackrel{+}\leftarrow - \triangledown_{\theta_{E_x}} (\mathcal{L_R}$)

\State $ \theta_{G_x}, \theta_{D_x} \stackrel{+}\leftarrow - \triangledown_{\theta_{G_x}, \theta_{D_x}} (\mathcal{L_D}$)

\State $ \theta_{G_z}, \theta_{D_c} \stackrel{+}\leftarrow - \triangledown_{\theta_{G_z}, \theta_{D_c}} (\mathcal{L_Z}$)

\EndWhile\label{aclgandwhile}
\end{algorithmic}

\end{algorithm}



\section{Experiments}
In this section, we present experimental results for each of the above discussed architectures compared with both classic and the recent advanced resident models. We carry out the comparison on several benchmark datasets inlcluding MNIST, CIFAR10, and CelabA. We shall also use synthetic data for a brief discussion about the selection of prior noise impacted to our ACL module.

As a proof of concept, we implement our networks with minimum optimization. Specifically, we use simple 2-layer MLPs in the ACL module. All the experiments, except where noted, are carried out with the same experimental settings for fair comparison.

\subsection{Datasets}
In this study, we experimented on three datasets,namely MNIST \cite{lecun1998gradient}, CelebA \cite{liu2015faceattributes}, and CIFAR-10\cite{krizhevsky2009learning}. The MNIST dataset is a well-known image dataset, with 70,000 28$\times$28 handwritten-digit images. The CelebA dataset contains 202,599 celebrity images of varying resolutions. The CIFAR-10 dataset is a collection of 60,000 32$\times$32 images covering a wide range of objects. 

\subsection{Comparison with Adversarial Autoencoders} \label{exp-comparison-gans}
\begin{figure}[htb]
\begin{center}
  \includegraphics[height=0.42\linewidth,width=0.45\linewidth]{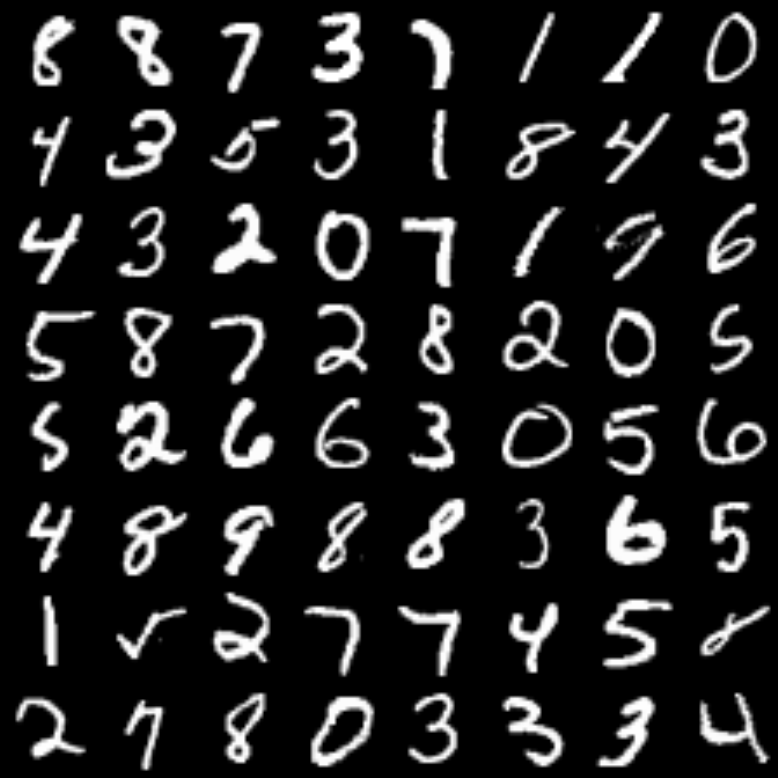}
  \vspace{.2em}
  \includegraphics[height=0.42\linewidth,width=0.45\linewidth]{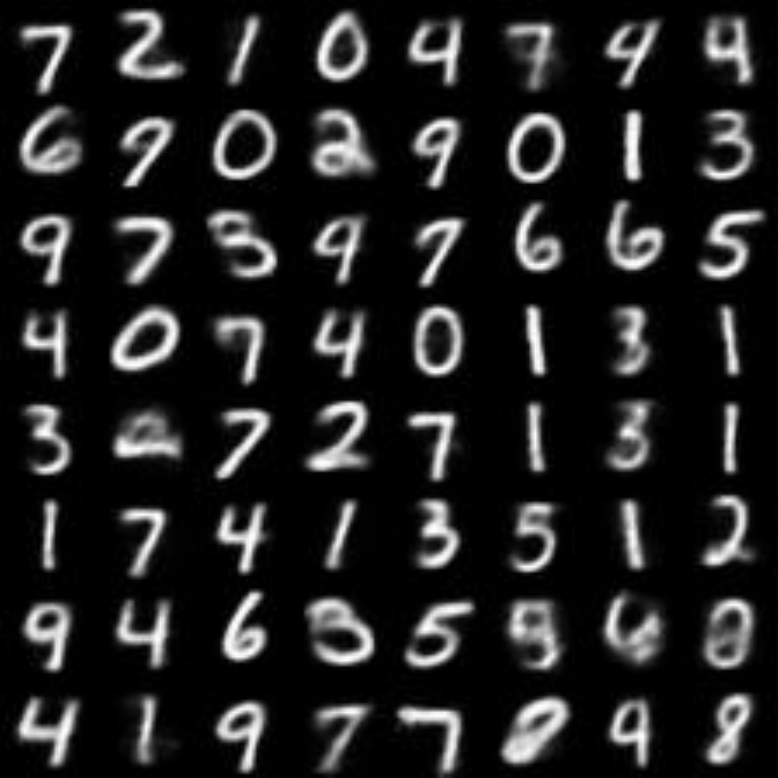}
  \vspace{.2em}
  \includegraphics[height=0.42\linewidth,width=0.45\linewidth]{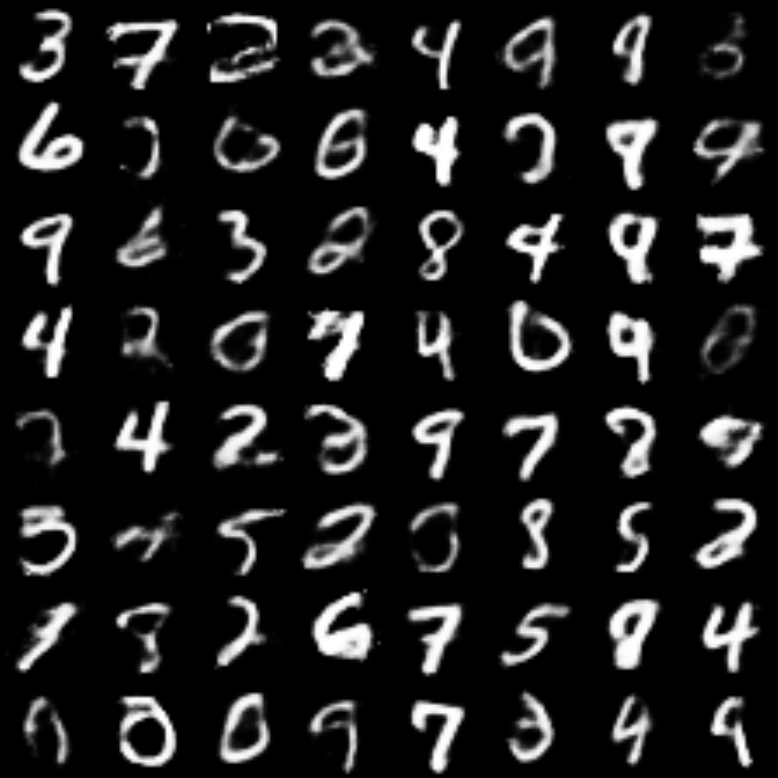}
  \includegraphics[height=0.42\linewidth,width=0.45\linewidth]{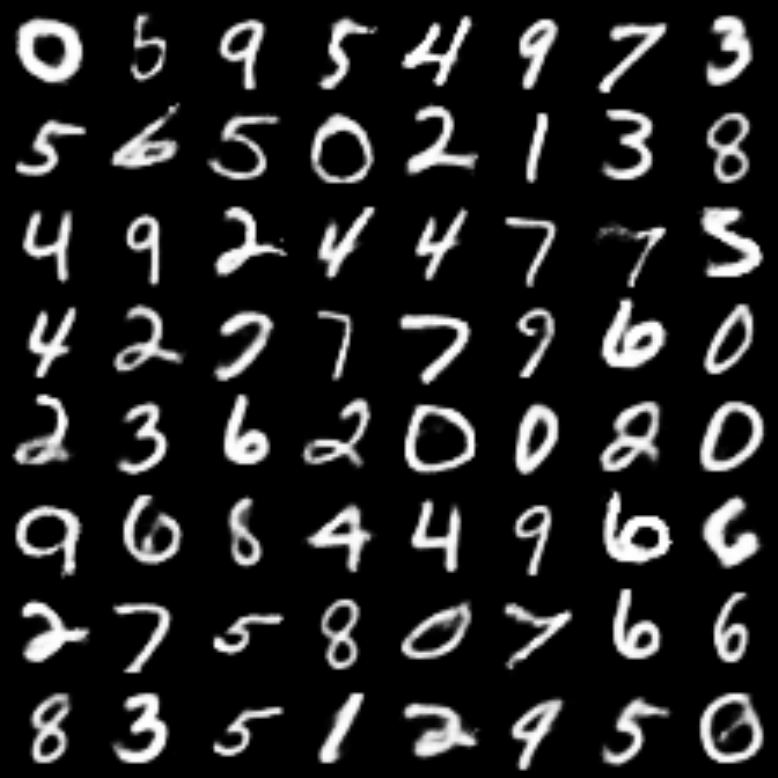}
\end{center}
\vspace{-1.5em}
\caption{\textbf{Top left}: Real images of MNIST; \textbf{Top right}: VAE generated output; \textbf{Bottom left}: AAE output; \textbf{Bottom right}: Our ACL-AE results. Note that, all the synthetic results are generated from random sampled noise data.}
\label{fig:ae_mnist}
\end{figure}

To examine whether our ACL model can improve existing Autoencoder models, we compare our ACL-AE with VAE and AAE on MNIST dataset for image generation task. We present our experimental results in Figure \ref{fig:ae_mnist}. The results are from one mini-batch after 100 epochs with dimension=10 in the latent space. The three experiments follow the same settings in \footnote{https://github.com/hwalsuklee/tensorflow-mnist-AAE}. From the generated images, we can see that the images generated from VAE are usually smooth but more vague, while the images generated from AAE are sharper but severe distortions are easier to be found comparing to other models. Our method is able to generate clear, smooth and realistic samples compared to the real images.

The AAE model tries to regularize the latent code learning by a random noise distribution, which may partially explain the distortion of shapes, since the regularization may contradict to the image construction. On the contrary, our ACL model regularizes the generated code distribution with the latent code from Autoencoders. Another comparison on CIFAR-10 further demonstrates our analysis. The evaluation metric we used is the Fréchet Inception Distance (FID) score introduced in previous work \cite{heusel2017gans}. FID score is proved to be able to capture the similarity of generated images to real ones better than Inception Score \cite{salimans2016improved}. On CIFAR-10, our ACL-AE model outperforms AAE by a large margin that can be seen in Figure \ref{fig:acl_vs_aae} left.

\subsection{Comparison with GANs}

In this section, we show the performance of our ACL-GAN model with comparisons to both Autoencoder regularized GANs and conventional GANs. 

\begin{figure}[htb]
\begin{center}
  \includegraphics[trim={6cm 18cm 6cm 3.2cm}, clip, width=0.49\linewidth]{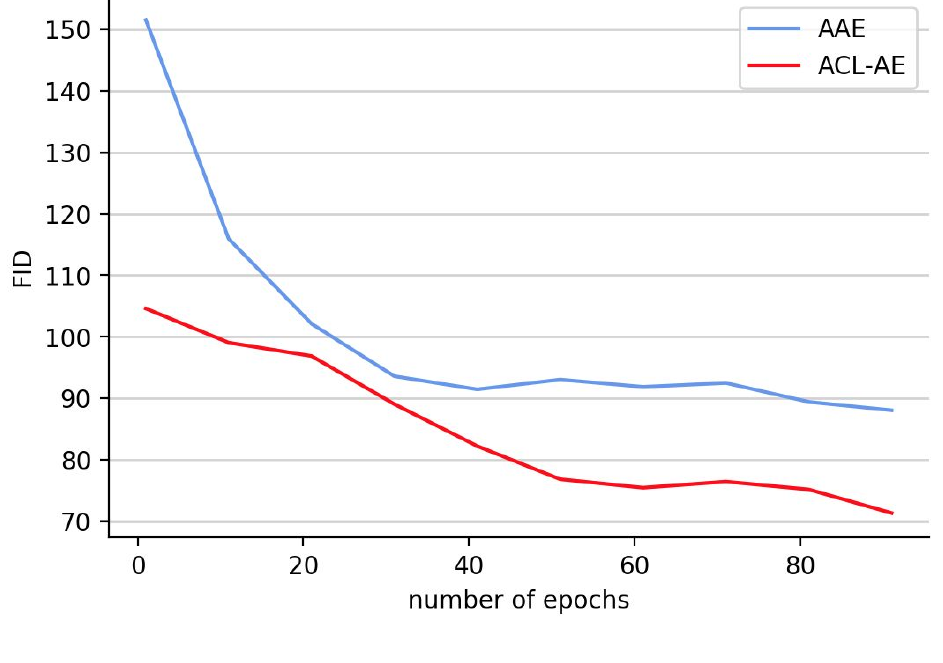}
  \includegraphics[trim={6cm 18cm 6cm 3.2cm}, clip, width=0.49\linewidth]{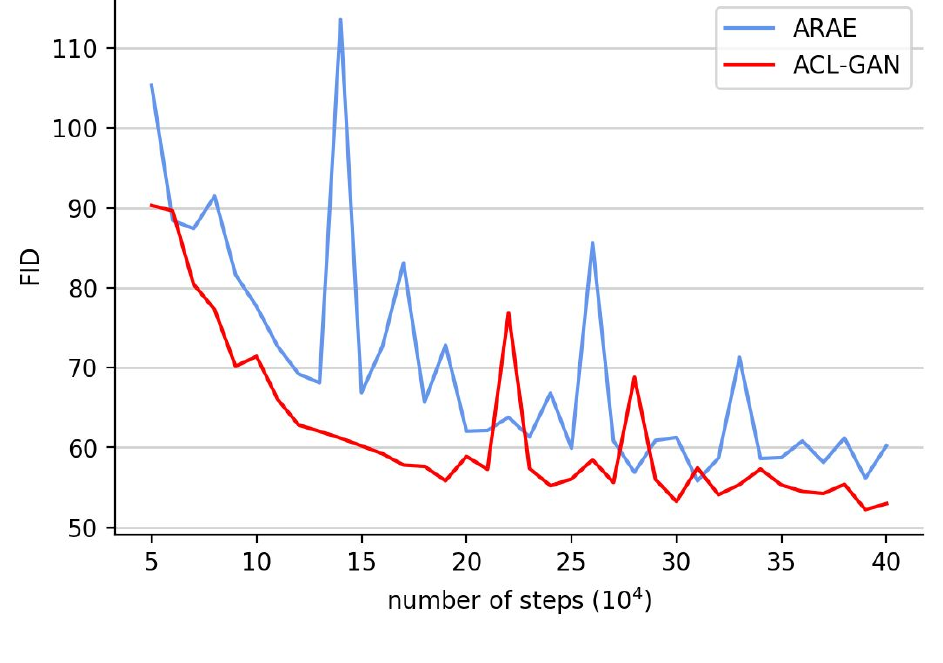}
\end{center}
\vspace{-1.5em}
\caption{FID scores on CIFAR-10. \textbf{Left}: ACL-AE vs AAE; \textbf{Right}: ACL-GAN vs ARAE-GAN.}
\label{fig:acl_vs_aae}
\end{figure}

Our ACL-GAN model extends Autoencoders to be jointly trained with an image discriminator. There are other works sharing the same spirit, e.g., adversarially regularized autoencoders (ARAE)\cite{junbo2017arae}. These models become a new family that is between Autoencoders and conventional GANs. The model ARAE and our ACL-GAN are similar in adversarial learning between the prior distribution $p_{z}(\bm{z})$ and the latent code distribution $p_{c}(\bm{c})$. However, they are leading to very different results by their underlying assumptions and designs. The major difference is that ACLs map $p_{c}(\bm{c})$ to “target” $p_{c}(\bm{c})$ while ARAE mutually aligns the two distributions during the learning. We performed experiments on CIFAR-10 to better support our discussion. One can observe that our ACL-GAN model outperforms ARAE in Figure \ref{fig:acl_vs_aae} right in terms of FID scores.

We further compare our ACL-GAN with both vanilla GAN and WGAN-GP \cite{gulrajani2017improved}, the latter one is considered as a state-of-the-art method. For the experimental settings of GAN and WGAN-GP, we follow the original specifications in their published papers and code repositories. Our ACL-GAN and ACL-GAN-GP also follow the same settings for fair comparison. Specifically, for ACL-GAN, we simply attach our ACL module to vanilla GAN, keeping everything else unchanged. For ACL-GAN-GP, we apply the gradient penalty \cite{gulrajani2017improved} for comparison. 

\begin{figure}[htb]
\begin{center}
\includegraphics[width=0.49\linewidth]{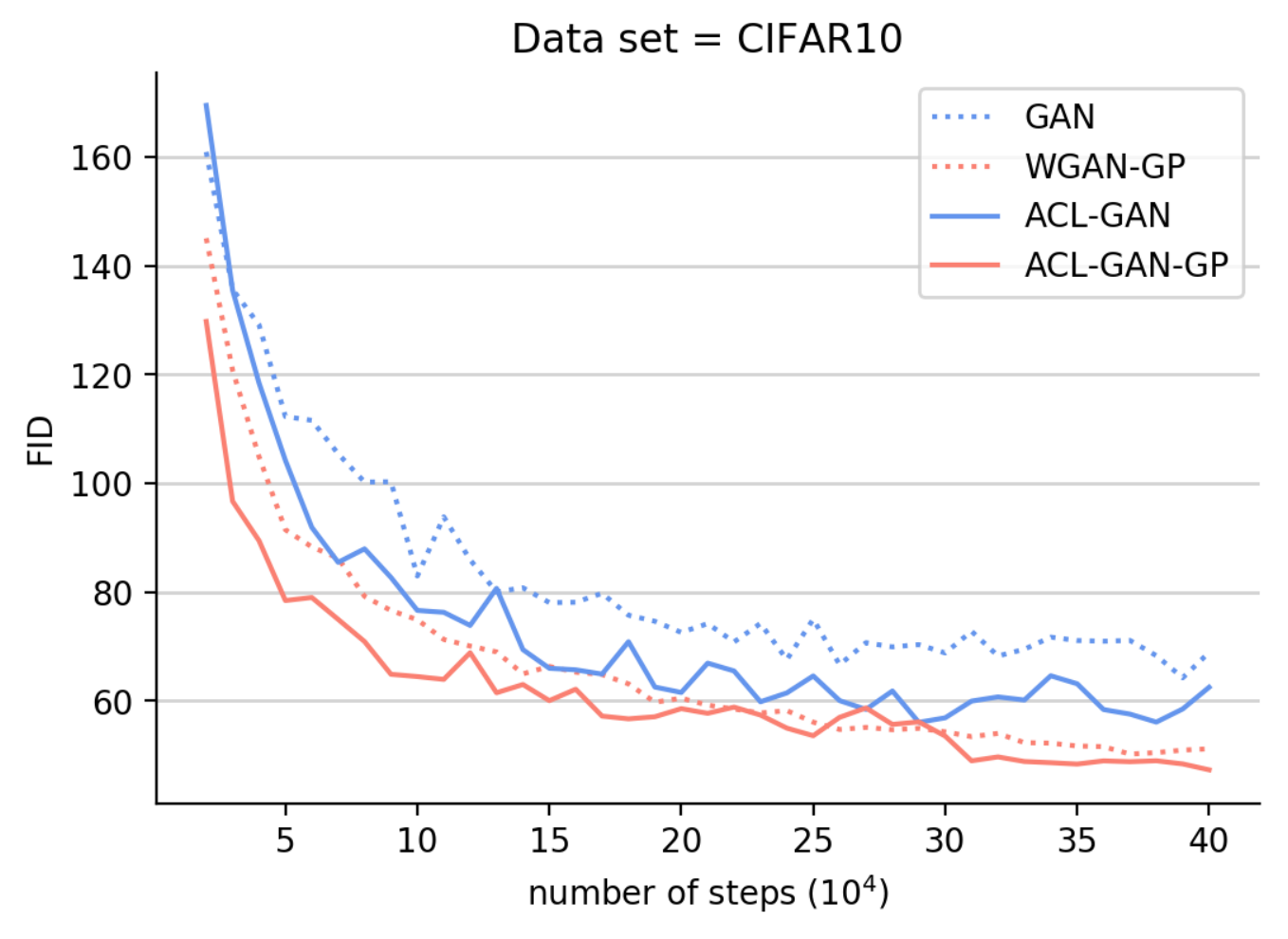}
\includegraphics[width=0.49\linewidth]{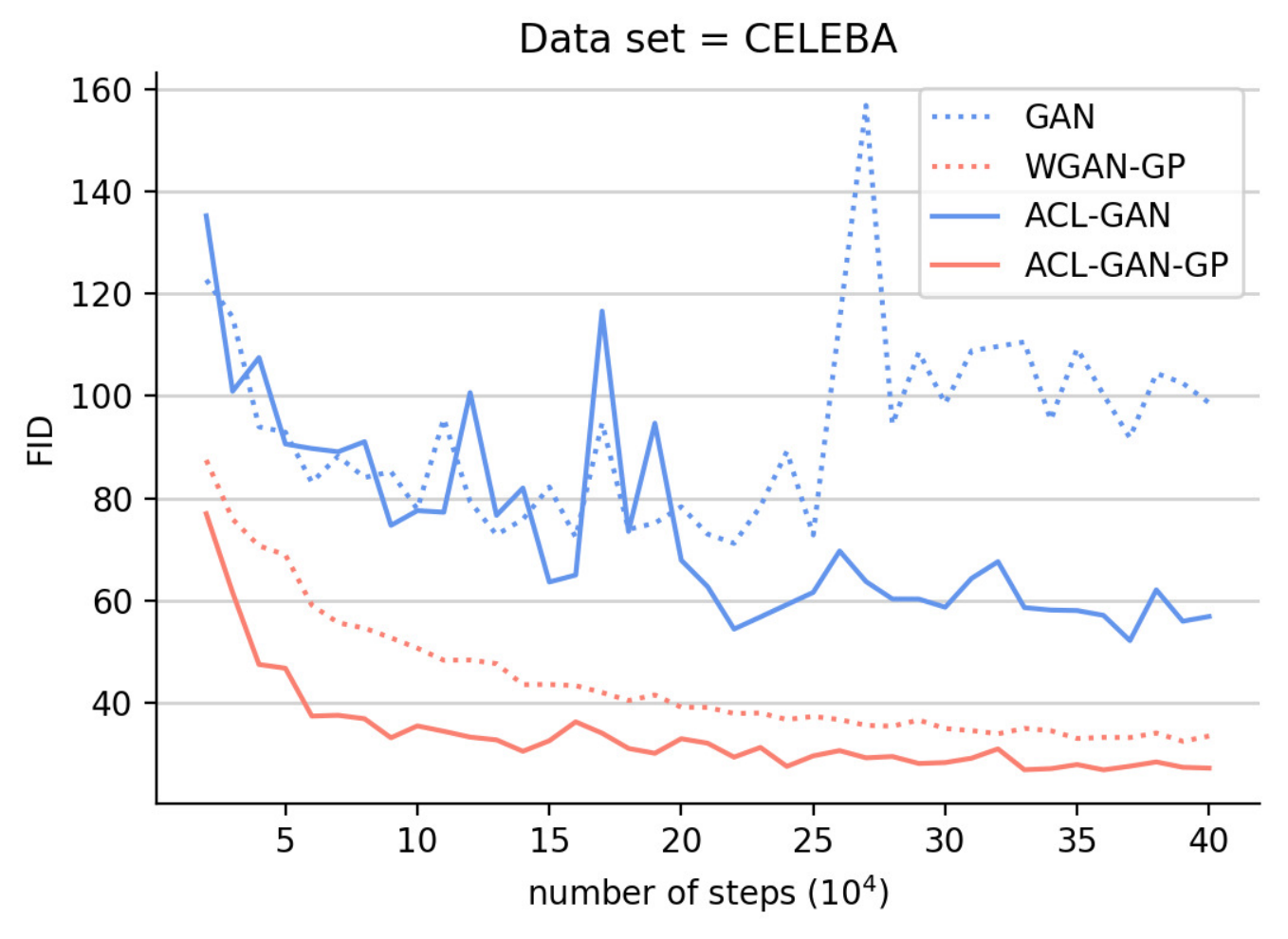}
\caption{The FID scores of GAN, ACL-GAN, WGAN-GP, and ACL-GAN-GP on the CIFAR10 and CelebA dataset in a single training process.}
\label{fig:acl_gan_celeba_cifar10}
\end{center}
\end{figure}

We report the FID scores in table \ref{tab:fid_scores} following the experimental setup in\cite{lucic2017gans}. The big margin improvements demonstrated the advantage of the integration of our ACL module to the original GAN and the WGAN-GP architectures. We also plot the FID score curves for each model in a single training pass for CELEBA and CIFAR-10 datasets separately (Figure \ref{fig:acl_gan_celeba_cifar10}). From the plots, we can observe that GAN models with Gradient Penalty (GP) loss generally perform better than vanilla GAN models (red versus blue lines). This can be observed consistently before and after plugging our ACL module. In addition, integration of our ACL module consistently improves model performances (solid versus dashed lines). We can conclude that with integration of very simple, e.g., two-layer perceptron ACL modules, one can significantly improve regular GANs' performance. 


\begin{table}[htb]
\begin{center}
\begin{tabular}{lcc}
\hline
\multirow{2}{*}{\textbf{Models}} &\multicolumn{2}{c}{\textbf{Datasets}}\\\cline{2-3}
                                     & \textbf{CIFAR-10} & \textbf{CelebA} \\
\hline
GAN       & 67.07 $\pm$ 2.82 & 64.28 $\pm$ 2.12 \\
\textbf{ACL-GAN}    & 56.05 $\pm$ 0.36 & 52.02 $\pm$ 0.25  \\
\hline
WGAN-GP    & 50.22 $\pm$ 0.74 & 32.85 $\pm$ 0.53 \\
\textbf{ACL-GAN-GP} & \textbf{47.32 $\pm$ 0.67} & \textbf{26.91 $\pm$ 0.22}\\
\hline
\end{tabular}
\end{center}
\vspace{-1.5em}
\caption{The FID score comparison on CIFAR-10 and CelebA of different models. ACL models are our versions.}
\label{tab:fid_scores}
\end{table}

\vspace{-2.5em}

\subsection{ACL-GANs with Supervision}
\begin{figure}[htb]
\begin{center}
\includegraphics[width=0.49\linewidth]{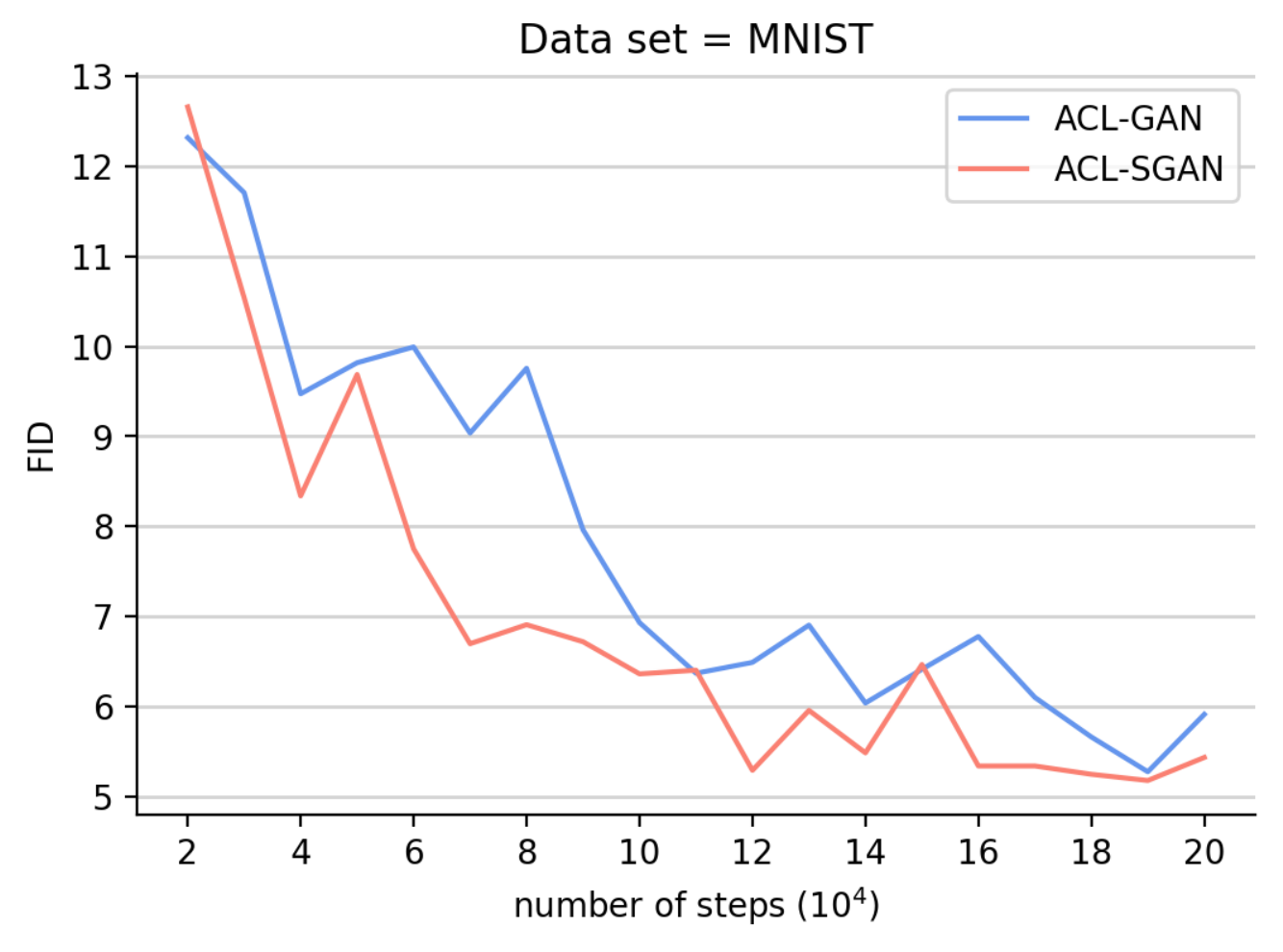}
\includegraphics[width=0.49\linewidth]{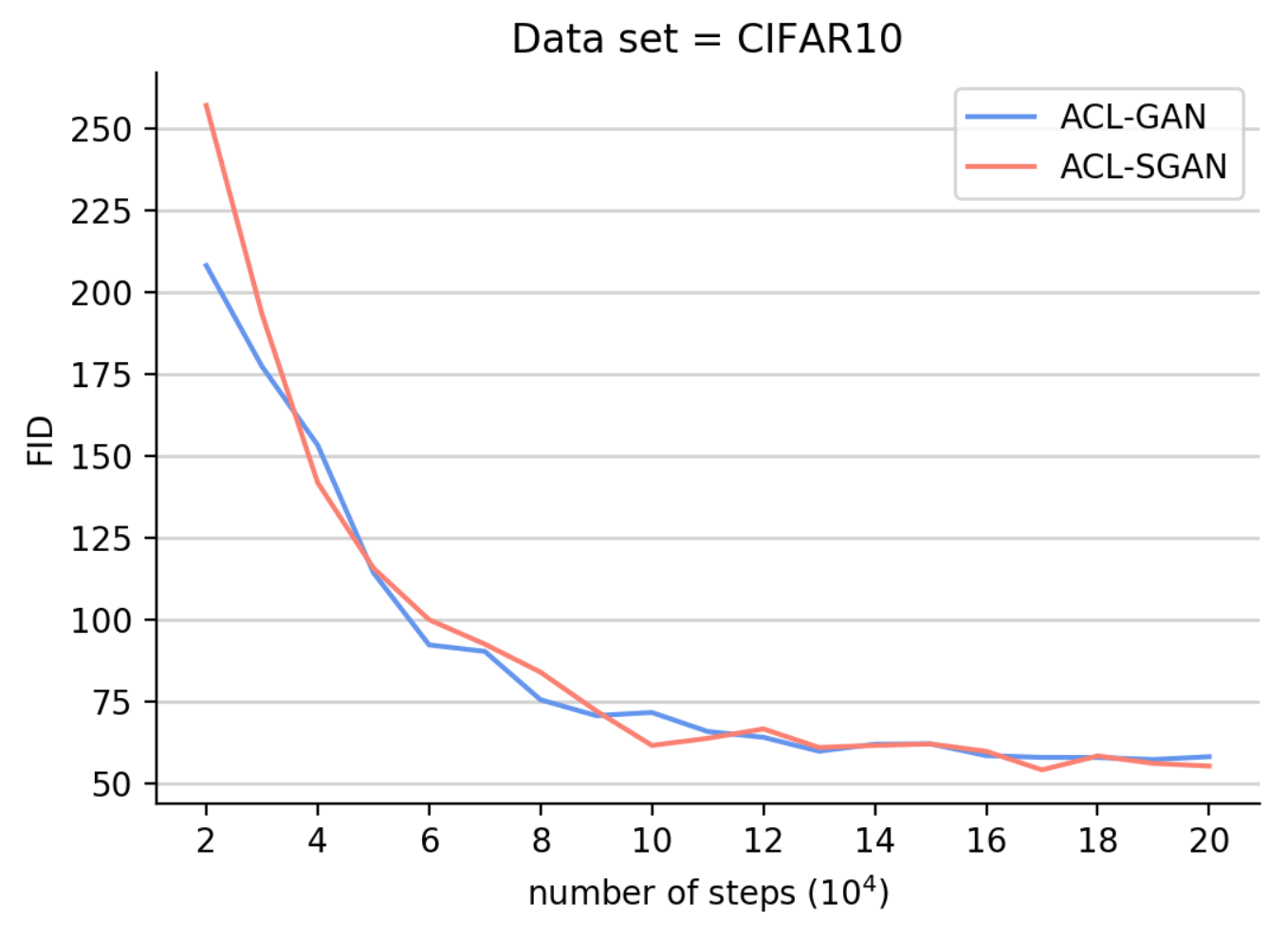}
\caption{The FID scores of ACL-GAN and ACL-SGAN on CIFAR10 and CelebA in a single training process.}
\label{fig:acl_gan_sgan}
\end{center}
\end{figure}

Joint training of a classification model with a GAN is non trivial in a common base \cite{dai2017good}. It is also very interesting to investigate more in this joint model since its potential to both semi-supervised learning and image generation. However, we lack deep analysis due to time constraints and limited resources. Thus we keep our focus on the discussion of our ACL module on generative modeling tasks. From Figure \ref{fig:acl_gan_sgan}, it is pleased to see that by using a classic classifier as our inference net, the overall generative modeling performance is not degraded, or even slightly better than our unsupervised version. 

\vspace{-1.5em}

\subsection{Data Augmentation}

To examine how well ACL helps to generalize the mapping to unseen regions in the latent code space, we also present our experimental results on data augmentation task. Given two data points, through linear interpolation between the two samples in the latent space, we can generate massive believable yet unseen samples from existing datasets. The generated data, when of high quality, may be very beneficial for various computer vision tasks where there is hunger for more training data.  

\begin{figure}[htb]
\begin{center}
  \includegraphics[width=1\linewidth]{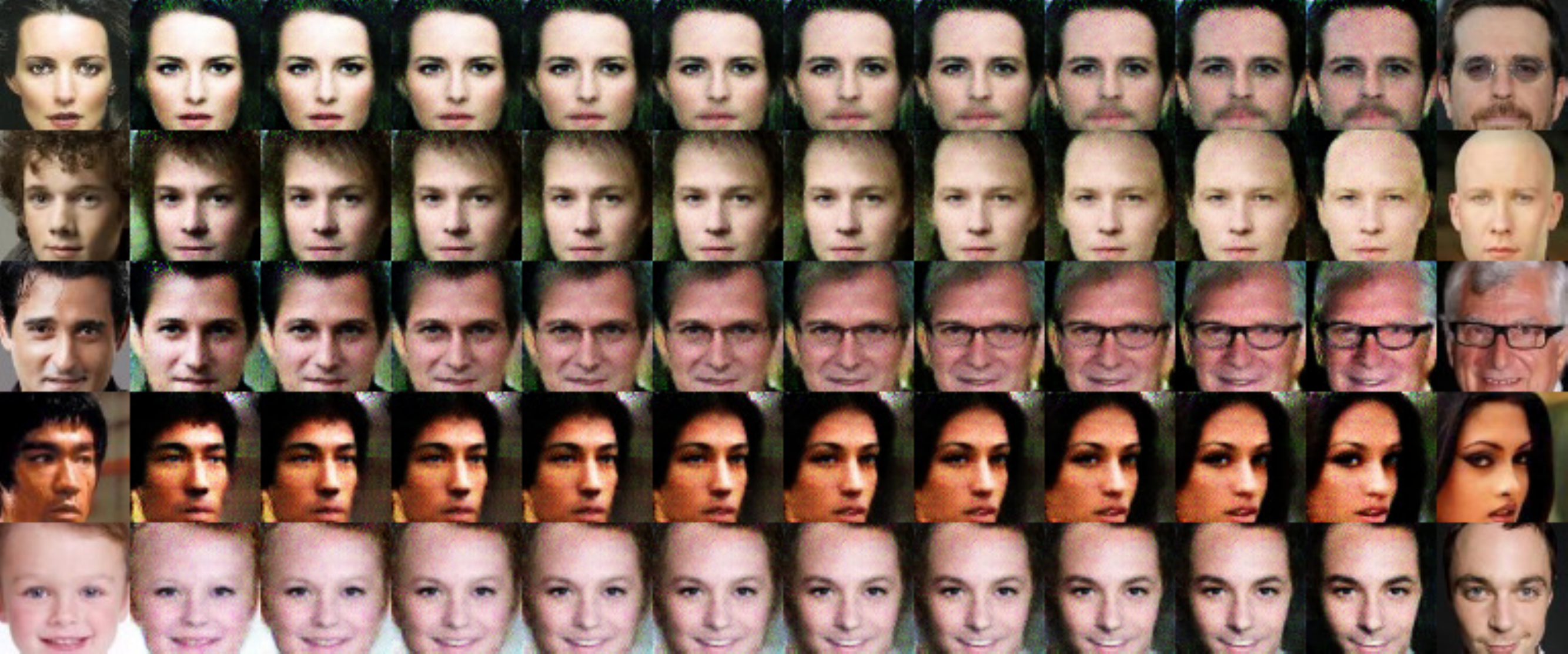}
\end{center}
\vspace{-1.5em}
\caption{The interpolation results: the leftmost and rightmost images are real samples, and images in between are generated from interpolated latent points.}
\label{fig:acl_aug}
\end{figure}

\begin{figure*}[htb]
\begin{center}

\begin{subfigure}{.8\textwidth}
\begin{minipage}{0.15\textwidth}
\includegraphics[trim={0cm 1cm 0cm 1cm}, clip, width=1\linewidth]{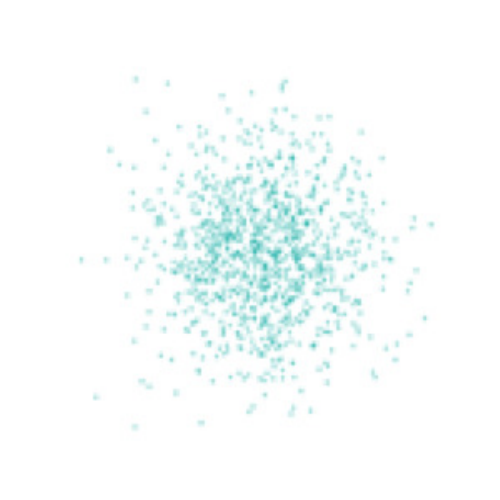}
\end{minipage}
\begin{minipage}{0.69\textwidth}
\includegraphics[trim={0.5cm 1cm 0 0.5cm}, clip, width=1\linewidth]{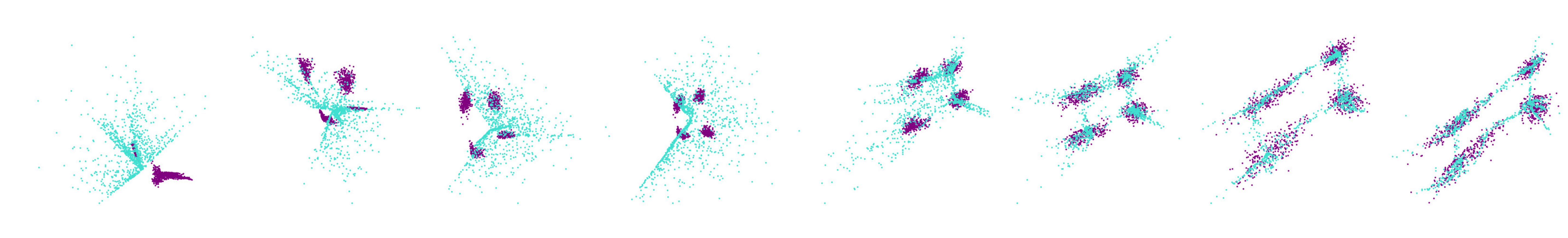}
\includegraphics[trim={0.5cm 0.5cm 0 1cm}, clip, width=1\linewidth]{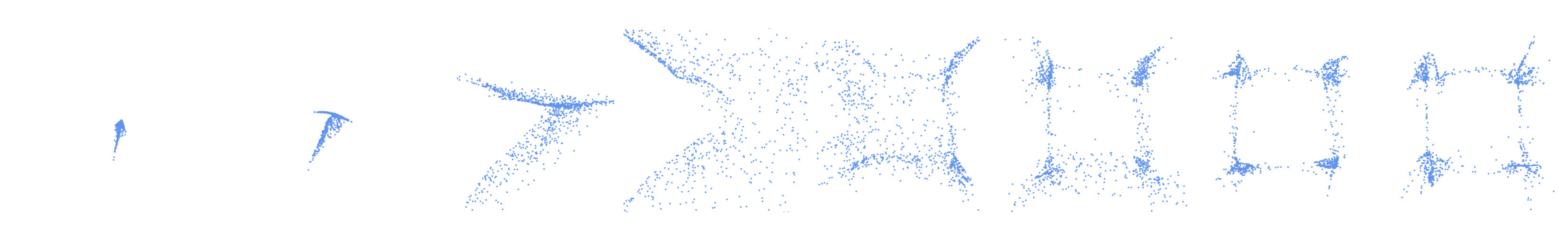}
\end{minipage}
\begin{minipage}{0.15\textwidth}
\includegraphics[trim={0cm 0.5cm 0cm 0.5cm}, clip, width=1\linewidth]{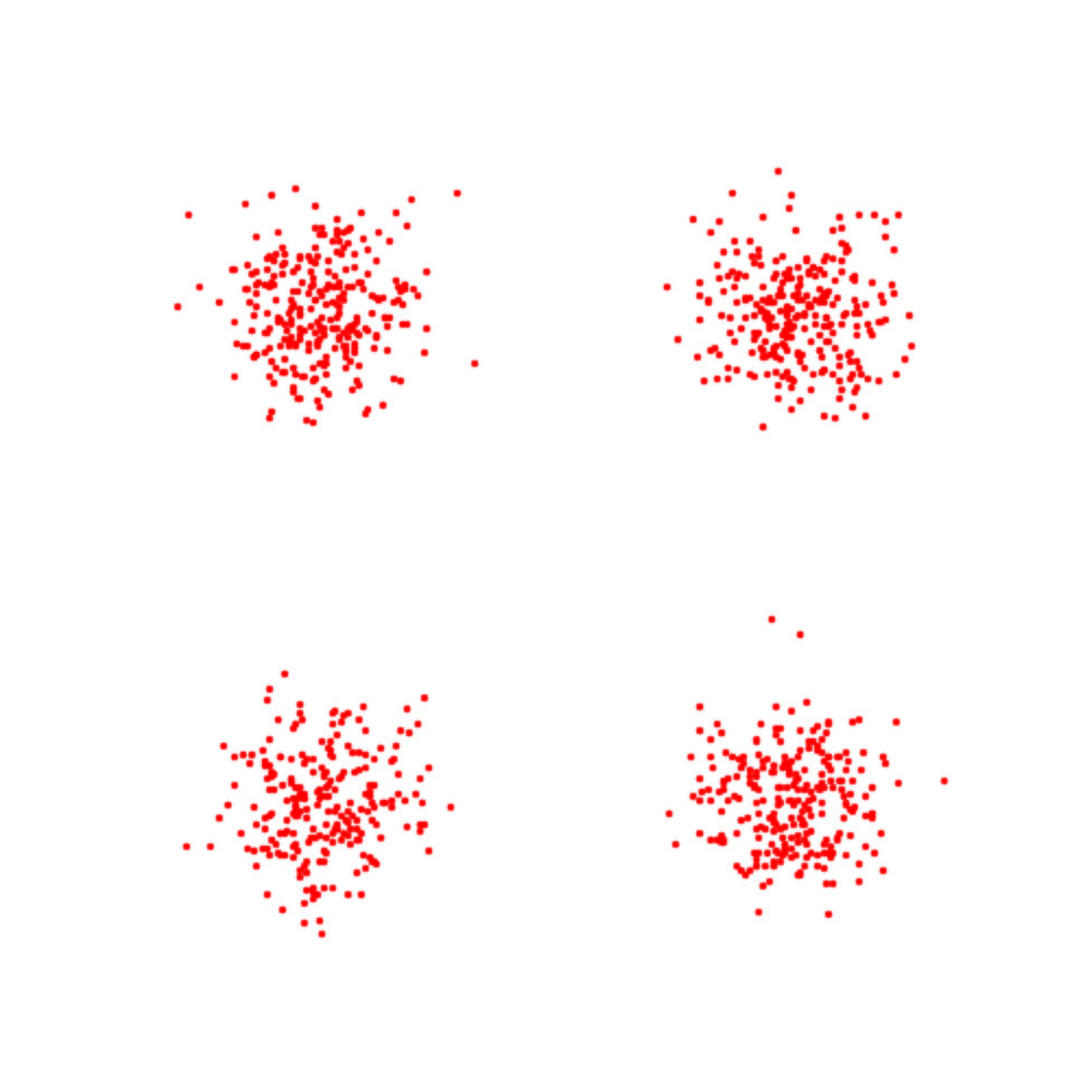}
\end{minipage}
\caption{}
\end{subfigure}

\begin{subfigure}{.8\textwidth}
\begin{minipage}{0.15\textwidth}
\includegraphics[trim={0cm 1cm 0cm 1cm}, clip, width=1\linewidth]{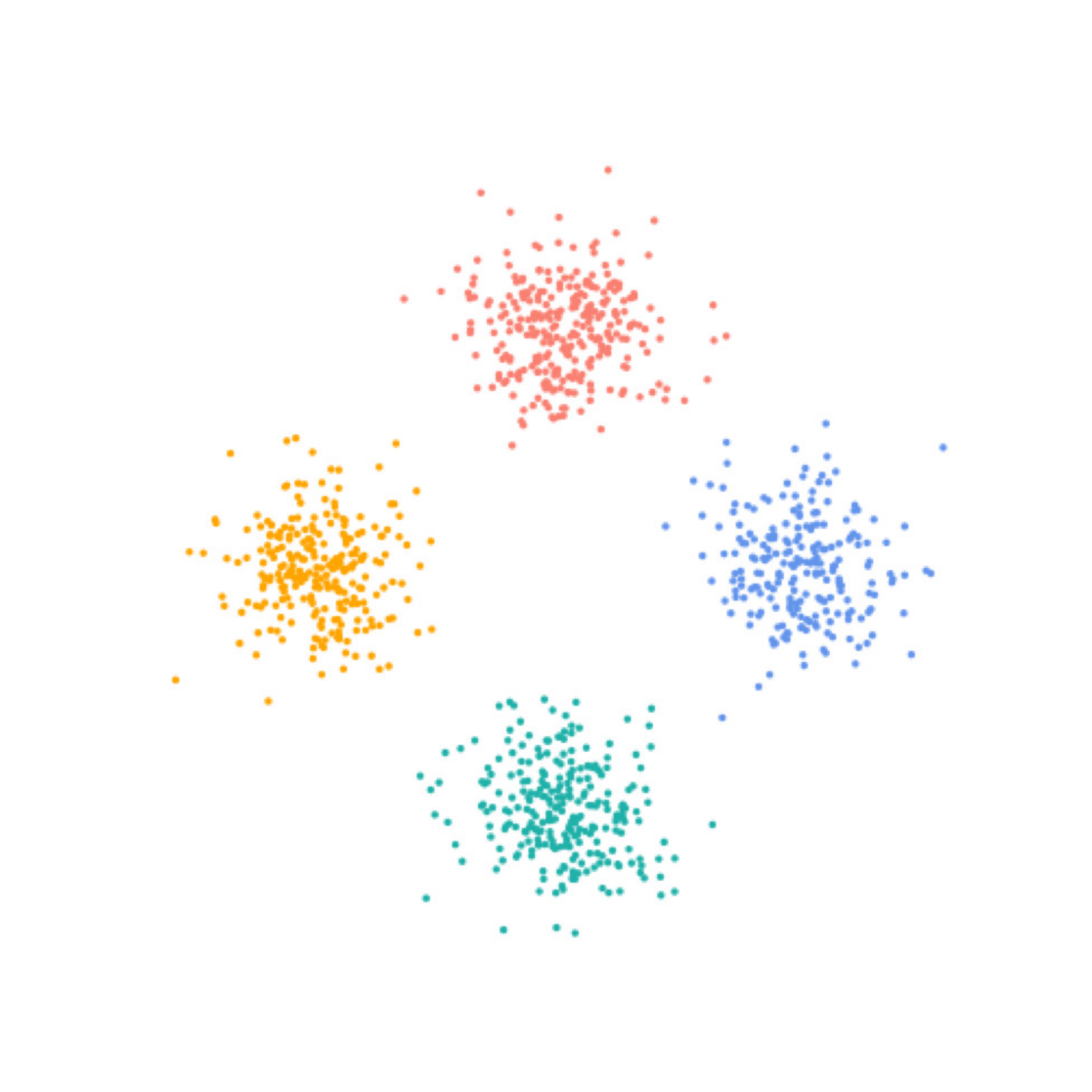}
\end{minipage}
\begin{minipage}{0.69\textwidth}
\includegraphics[trim={0.5cm 1cm 0 0.5cm}, clip, width=1\linewidth]{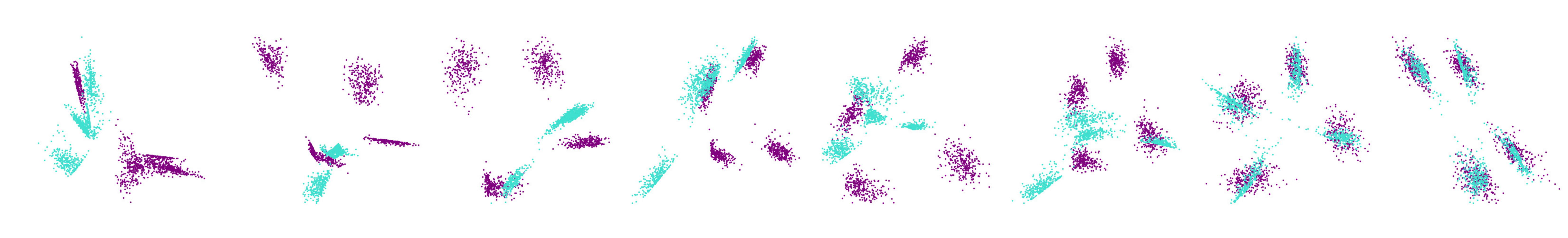}
\includegraphics[trim={0.5cm 0.5cm 0 1cm}, clip, width=1\linewidth]{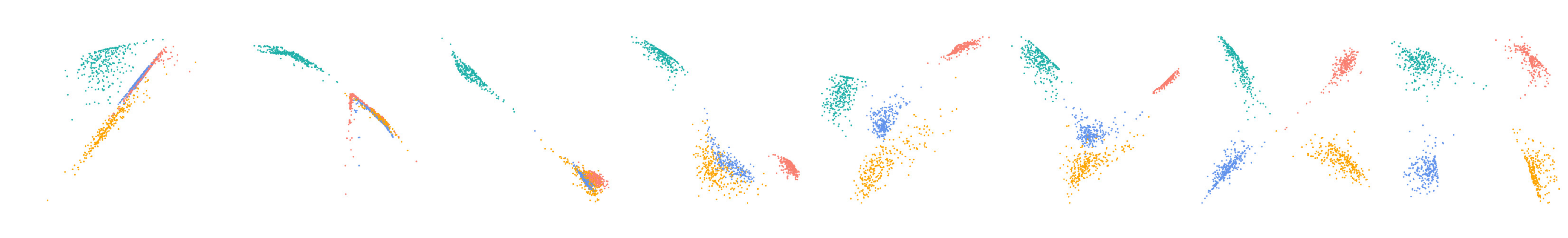}
\end{minipage}
\begin{minipage}{0.15\textwidth}
\includegraphics[trim={0cm 0.5cm 0cm 0.5cm}, clip, width=1\linewidth]{fig/real.pdf}
\end{minipage}
\caption{}
\end{subfigure}

\end{center}
\vspace{-1.5em}
\caption{Illustration of piror impact to the ACL learning. There are two toy examples (a) and (b). While sharing a same learning target, i.e., a 4-mode Gaussian distribution (on rightmost), setting (a) takes a 1-mode Gaussian Prior and (b) uses a 4 mode Gaussian Prior (on leftmost). respectively. The middles show the real code distribution (in purple) from the inference net, and the ACL learned fake distributions (in turquoise), in different learning steps.}
\label{fig:prior}
\end{figure*}

The interpolation results are shown in figure \ref{fig:acl_aug}. The images on the leftmost and rightmost ends of each row are real images from the CELEBA dataset. The images in between are generated from the interpolated latent codes. We have two simple observations. First, all generated samples are of high quality. Second, the transitions from one sample to the next are consistently smooth. This can be taken as a sanity check for overfitting, where our ACL models seems to be able to generate samples not limited to regions near training data in the latent space, thus exhibiting good generalization ability. Different from the gradually-changed generated images with noise grid, our interpolated images are converting from one real sample to another. This demonstrates that ACL is not limited in generating images of high quality, but shows promising capability in controlling the generated samples through manipulating the input from the latent space.

\subsection{Prior Analysis}
 We have shown the power of ACL in different modeling frameworks with a simple single-mode input, e.g., 1-mode Gaussian distribution. This section moves one-step further to discuss how the prior selection of ACL will change its learning behaviors.

 We carry out a toy experiment with ACL-AE using synthetic data. The target distribution is a 4-mode two-dimensional Gaussian, plotted as four red clusters in Figure \ref{fig:prior}. For the prior distribution, we experimented two different settings. Random noise is sampled from one and four two-dimensional Gaussian modes for setting one and two, as shown in Figure \ref{fig:prior}. We shall examine whether adjusting the prior can be helpful for generative tasks. Since in Setting 2, the prior shares a same distribution to the target statistic, we are expecting to see some advantage over Setting 1.
 
 
 We have several observations from this visualization: 1) In both 1-mode and 4-mode Gaussian priors, the learned codes by ACL models (in turquoise) are progressively better aligned with the latent codes from the autoencoders (in purple) throughout the training steps; 2) The distribution of the generated data distribution is closely related to the alignment between the learned codes and the latent codes of autoencoders; 3) The final generated data distribution is closer to the target distribution when the prior is switched to the 4-mode Gaussian \ref{fig:prior}(b) than the 1-mode setting \ref{fig:prior}(a). 
 
 The conclusion of this toy experiment supports our speculation that a proper prior selection can be a way of improving the code learning, thus consequently benefiting the final data generation. How to find proper prior distributions for real applications seems to be an interesting and challenging problem.

\section{Conclusions and Future Work}
In this paper, we propose an adversarial code learning(ACL) framework, which jointly trains an inference network, an adversarial code generative net, and an adversarial image generative net. The ACL module generalizes autoencoders to be generative models, and boosts up several popular GAN structures' performance. Even though ACLs fall back to the min-max problem, we move the adversary from an unknown and more complex pixel space to much lower dimensional latent spaces, e.g., a single mode Gaussian prior. The experimental results demonstrated the effectiveness.

This framework admits other straightforward extensions. For example, with a learning target to a late intermediate layer in a classification model, ACL smoothly converts a classic discriminative model to a generative model. The joint model can be used in semi-supervised learning, which needs to be further investigated. Extending to other sophisticated models is an analogy to the presented models, as long as a latent code distribution is given, demonstrated the potential usage in many applications.

{\small
\bibliographystyle{ieee}
\bibliography{egbib}
}

\end{document}